\documentclass[11pt,a4paper]{article}
\usepackage[hyperref]{emnlp-ijcnlp-2019}
\usepackage{times}
\usepackage{latexsym}
\usepackage{graphicx}
\usepackage{amsfonts}

\usepackage{amsmath}
\usepackage{algorithm}
\usepackage{algorithmic}

\usepackage{wasysym}
\usepackage{hyperref}
\usepackage{verbatim}

\usepackage{booktabs}
\usepackage{multirow}
\usepackage{url}

\usepackage{soul}
\setuldepth{c}

\usepackage{array}
\usepackage{comment}
\usepackage{makecell}

\usepackage[colorinlistoftodos, prependcaption]{todonotes}

\aclfinalcopy 

\title{PubMedQA: A Dataset for Biomedical Research Question Answering}

\author{Qiao Jin \\
  University of Pittsburgh \\
  {\tt qiao.jin@pitt.edu} \\\And
  Bhuwan Dhingra \\
  Carnegie Mellon University \\
  {\tt bdhingra@cs.cmu.edu} \\\And
  Zhengping Liu \\
  University of Pittsburgh \\
  {\tt zliu@pitt.edu} \\
  \AND
  William W. Cohen \\
  Google AI \\
  {\tt wcohen@google.com} \\\And
  Xinghua Lu \\
  University of Pittsburgh \\
  {\tt xinghua@pitt.edu} \\
  }

\date{}

\begin{document}

\setlength{\abovedisplayskip}{8pt}
\setlength{\belowdisplayskip}{8pt}

\maketitle
\begin{abstract}
We introduce PubMedQA, a novel biomedical question answering (QA) dataset collected from PubMed abstracts. The task of PubMedQA is to answer research questions with yes/no/maybe 
(e.g.: \textit{Do preoperative statins reduce atrial fibrillation after coronary artery bypass grafting?})
using the corresponding abstracts. PubMedQA has 1k expert-annotated, 61.2k unlabeled and 211.3k artificially generated QA instances. Each PubMedQA instance is composed of (1) a question which is either an existing research article title or derived from one, (2) a context which is the corresponding abstract without its conclusion, (3) a long answer, which is the conclusion of the abstract and, presumably, answers the research question, and (4) a yes/no/maybe answer which summarizes the conclusion. PubMedQA is the first QA dataset where reasoning over biomedical research texts, especially their quantitative contents, is required to answer the questions. Our best performing model, multi-phase fine-tuning of BioBERT with long answer bag-of-word statistics as additional supervision, achieves 68.1\% accuracy, compared to single human performance of 78.0\% accuracy and majority-baseline of 55.2\% accuracy, leaving much room for improvement. PubMedQA is publicly available at \url{https://pubmedqa.github.io}.
\end{abstract}

\section{Introduction}
\begin{figure}[!t]
    \framebox{
    \parbox{0.45\textwidth}{
    \small
    \textbf{\ul{Question}:} \newline
    Do preoperative statins reduce atrial fibrillation after coronary artery bypass grafting? \newline
    \textbf{\ul{Context}:} \newline
    \textit{\textbf{(Objective)}} Recent studies have demonstrated that statins have pleiotropic effects, including anti-inflammatory effects and atrial fibrillation (AF) preventive effects [...] \newline
    \textit{\textbf{(Methods)}} 221 patients underwent CABG in our hospital from 2004 to 2007. 14 patients with preoperative AF and 4 patients with concomitant valve surgery [...] \newline
    \textit{\textbf{(Results)}} The overall incidence of postoperative AF was 26\%. \textcolor{blue}{\textit{\textbf{Postoperative AF was significantly lower in the Statin group compared with the Non-statin group (16\% versus 33\%, p=0.005).}}} Multivariate analysis demonstrated that  independent predictors of AF [...] \newline
    \textbf{\ul{Long Answer}:} \newline
    \textit{\textbf{(Conclusion)}} Our study indicated that preoperative statin therapy seems to reduce AF development after CABG. \newline
    \textbf{\ul{Answer}:} yes
    }
    }
    \vskip -0.15cm
    \caption{An instance \cite{sakamoto2011preoperative} of PubMedQA dataset: Question is the original question title; Context includes the structured abstract except its conclusive part, which serves as the Long Answer; Human experts annotated the Answer yes. Supporting fact for the answer is \textcolor{blue}{\textit{\textbf{highlighted}}}.} 
    \label{fig:example}
    \vspace{-0.5em}
\end{figure}

A long-term goal of natural language understanding is to build intelligent systems that can reason and infer over natural language. The question answering (QA) task, in which models learn how to answer questions, is often used as a benchmark for quantitatively measuring the reasoning and inferring abilities of such intelligent systems.

While many large-scale annotated general domain QA datasets have been introduced \cite{rajpurkar2016squad, lai2017race, kovcisky2018narrativeqa, yang2018hotpotqa, kwiatkowski2019natural}, the largest annotated biomedical QA dataset, BioASQ \cite{tsatsaronis2015overview} has less than 3k training instances, most of which are simple factual questions. Some works proposed automatically constructed biomedical QA datasets \cite{pampari2018emrqa, pappas2018bioread, kim2018pilot}, which have much larger sizes. However, questions of these datasets are mostly factoid, whose answers can be extracted in the contexts without much reasoning.

In this paper, we aim at building a biomedical QA dataset which (1) has substantial instances with some expert annotations and (2) requires reasoning over the contexts to answer the questions. For this, we turn to the PubMed\footnote{\url{https://www.ncbi.nlm.nih.gov/pubmed/}}, a search engine providing access to over 25 million references of biomedical articles. We found that around 760k articles in PubMed use questions as their titles. Among them, the abstracts of about 120k articles are written in a structured style -- meaning they have subsections of ``Introduction'', ``Results'' etc. Conclusive parts of the abstracts, often in ``Conclusions'', are the authors' answers to the question title. Other abstract parts can be viewed as the contexts for giving such answers. This pattern perfectly fits the scheme of QA, but modeling it as abstractive QA, where models learn to generate the conclusions, will result in an extremely hard task due to the variability of writing styles.

Interestingly, more than half of the question titles of PubMed articles can be briefly answered by yes/no/maybe, which is significantly higher than the proportions of such questions in other datasets, e.g.: just 1\% in Natural Questions \cite{kwiatkowski2019natural} and 6\% in HotpotQA \cite{yang2018hotpotqa}. Instead of using conclusions to answer the questions, we explore answering them with yes/no/maybe and treat the conclusions as a long answer for additional supervision.

To this end, we present PubMedQA, a biomedical QA dataset for answering research questions using yes/no/maybe. We collected all PubMed articles with question titles, and manually labeled 1k of them for cross-validation and testing. An example is shown in Fig. \ref{fig:example}. The rest of yes/no/answerable QA instances compose of the unlabeled subset which can be used for semi-supervised learning. Further, we automatically convert statement titles of 211.3k PubMed articles to questions and label them with yes/no answers using a simple heuristic. These artificially generated instances can be used for pre-training. Unlike other QA datasets in which questions are asked by crowd-workers for existing contexts \cite{rajpurkar2016squad, yang2018hotpotqa, kovcisky2018narrativeqa}, in PubMedQA contexts are generated to answer the questions and both are written by the same authors. This consistency assures that contexts are perfectly related to the questions, thus making PubMedQA an ideal benchmark for testing scientific reasoning abilities.

As an attempt to solve PubMedQA and provide a strong baseline, we fine-tune BioBERT \cite{lee2019biobert} on different subsets in a multi-phase style with additional supervision of long answers. Though this model generates decent results and vastly outperforms other baselines, it's still much worse than the single-human performance, leaving significant room for future improvements.

\section{Related Works}
\paragraph{Biomedical QA:}
Expert-annotated biomedical QA datasets are limited by scale due to the difficulty of annotations. In 2006 and 2007, TREC\footnote{\url{https://trec.nist.gov/}} held QA challenges on genomics corpus \cite{hersh2006trec, hersh2007trec}, where the task is to retrieve relevant documents for 36 and 38 topic questions, respectively. QA4MRE \cite{penas2013qa4mre} included a QA task about Alzheimer's disease \cite{morante2012machine}. This dataset has 40 QA instances and the task is to answer a question related to a given document using one of five answer choices. The QA task of BioASQ \cite{tsatsaronis2015overview} has phases of (a) retrieve question-related documents and (b) using related documents as contexts to answer yes/no, factoid, list or summary questions. BioASQ 2019 has a training set of 2,747 QA instances and a test set of 500 instances.

Several large-scale automatically collected biomedical QA datasets have been introduced: emrQA \cite{pampari2018emrqa} is an extractive QA dataset for electronic medical records (EHR) built by re-purposing existing annotations on EHR corpora. BioRead \cite{pappas2018bioread} and BMKC \cite{kim2018pilot} both collect cloze-style QA instances by masking biomedical named entities in sentences of research articles and using other parts of the same article as context.

\paragraph{Yes/No QA:} Datasets such as HotpotQA \cite{yang2018hotpotqa}, Natural Questions \cite{kwiatkowski2019natural}, ShARC \cite{saeidi2018interpretation} and BioASQ \cite{tsatsaronis2015overview} contain yes/no questions as well as other types of questions. BoolQ \cite{clark2019boolq} specifically focuses on naturally occurring yes/no questions, and those questions are shown to be surprisingly difficult to answer. We add a ``maybe'' choice in PubMedQA to cover uncertain instances.

Typical neural approaches to answering yes/no questions involve encoding both the question and context, and decoding the encoding to a class output, which is similar to the well-studied natural language inference (NLI) task. Recent breakthroughs of pre-trained language models like ELMo \cite{peters2018deep} and BERT \cite{devlin2018bert} show significant performance improvements on NLI tasks. In this work, we use domain specific versions of them to set baseline performance on PubMedQA. 

\section{PubMedQA Dataset}
\subsection{Data Collection}
\begin{figure}
    \centering
    \includegraphics[width=\linewidth]{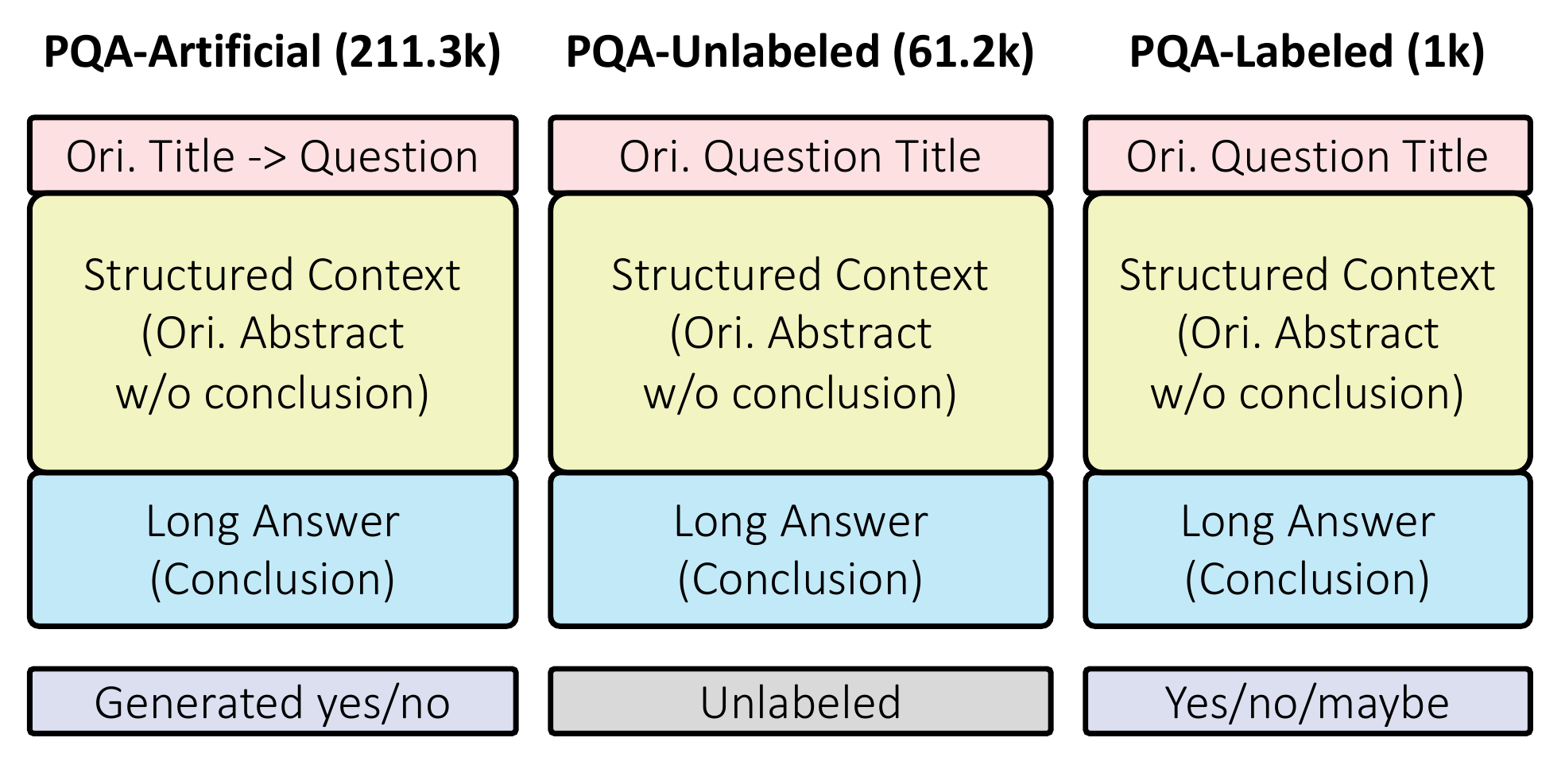}
    \vskip -0.15cm
    \caption{Architecture of PubMedQA dataset. PubMedQA is split into three subsets, PQA-A(rtificial), PQA-U(nlabeled) and PQA-L(abeled).}
    \label{fig:dataset}
    \vspace{-0.5em}
\end{figure}

PubMedQA is split into three subsets: labeled, unlabeled and artificially generated. They are denoted as PQA-L(abeled), PQA-U(nlabeled) and PQA-A(rtificial), respectively. We show the architecture of PubMedQA dataset in Fig. \ref{fig:dataset}.

\begin{table}[htbp]
\centering
\small
\begin{tabular}{lccc}
\toprule
\textbf{Statistic} & \textbf{PQA-L} & \textbf{PQA-U} & \textbf{PQA-A}\\
\midrule
Number of QA pairs & 1.0k & 61.2k & 211.3k\\
\midrule
Prop. of yes (\%) & 55.2 & -- & 92.8 \\
Prop. of no  (\%) & 33.8 & -- & 7.2 \\
Prop. of maybe (\%) & 11.0 & -- & 0.0 \\
\midrule
Avg. question length & 14.4 & 15.0 & 16.3 \\
Avg. context length & 238.9 & 237.3 & 238.0 \\
Avg. long answer length & 43.2 & 45.9 & 41.0 \\
\bottomrule
\end{tabular}
\vskip -0.15cm
\caption{PubMedQA dataset statistics.}
\vspace{-0.5em}
\label{tab:stat}
\end{table}

\paragraph{Collection of PQA-L and PQA-U:} PubMed articles which have i) a question mark in the titles and ii) a structured abstract with conclusive part are collected and denoted as pre-PQA-U. Now each instance has 1) a question which is the original title 2) a context which is the structured abstract without the conclusive part and 3) a long answer which is the conclusive part of the abstract.

Two annotators\footnote{Both are qualified M.D. candidates.} labeled 1k instances from pre-PQA-U with yes/no/maybe to build PQA-L using Algorithm \ref{algo:pqal}. The annotator 1 doesn't need to do much reasoning to annotate since the long answer is available. We denote this reasoning-free setting. However, the annotator 2 cannot use the long answer, so reasoning over the context is required for annotation. We denote such setting as reasoning-required setting.
Note that the annotation process might assign wrong labels when both annotator 1 and annotator 2 make a same mistake, but considering human performance in \S\ref{human}, such error rate could be as low as 1\%\footnote{Roughly half of the products of two annotator error rates.}. 500 randomly sampled PQA-L instances are used for 10-fold cross validation and the rest 500 instances consist of PubMedQA test set.

Further, we include the unlabeled instances in pre-PQA-U with yes/no/maybe answerable questions to build PQA-U. For this, we use a simple rule-based method which removes all questions started with interrogative words (i.e. wh-words) or involving selections from multiple entities. This results in over 93\% agreement with annotator 1 in identifying the questions that can be answered by yes/no/maybe.

\begin{algorithm}[t]
\small
    \caption{PQA-L data collection procedure}
    \label{algo:pqal}
\begin{algorithmic}
    \STATE {\bfseries Input: } pre-PQA-U
    \STATE $\text{ReasoningFreeAnnotation} \gets \{\}$
    \STATE $\text{ReasoningRequiredAnnotation} \gets \{\}$
    \STATE $\text{GroundTruthLabel} \gets \{\}$    
    \WHILE{not finished}
        \STATE Randomly sample an instance $inst$ from pre-PQA-U
        \IF {$inst$ is not yes/no/maybe answerable}
            \STATE Remove $inst$ and continue to next iteration
        \ENDIF
        \STATE Annotator 1 annotates $inst$  with $l_1 \in \{\text{yes}, \text{no}, \text{maybe}\}$ using question, context and long answer
        \STATE Annotator 2 annotates $inst$ with $l_2 \in \{\text{yes}, \text{no}, \text{maybe}\}$ using question and context
        \IF {$l_1 = l_2$}
            \STATE $l_a \gets l_1$
        \ELSE
            \STATE Annotator 1 and Annotator 2 discuss for an agreement annotation $l_a$
            \IF {\textbf{not} $\exists l_a$}
                \STATE Remove $inst$ and continue to next iteration
            \ENDIF
        \ENDIF
        \STATE $\text{ReasoningFreeAnnotation}[inst] \gets l_1$
        \STATE $\text{ReasoningRequiredAnnotation}[inst] \gets l_2$
        \STATE $\text{GroundTruthLabel}[inst] \gets l_a$
    \ENDWHILE
\end{algorithmic}
\end{algorithm}
\vspace{-0.5em}

\begin{table*}[!t]
\centering
\small
\begin{tabular}{p{6.5cm}p{6.5cm}cc}
\toprule
\textbf{Original Statement Title} & \textbf{Converted Question} & \textbf{Label} & \textbf{\%}\\
\midrule
Spontaneous electrocardiogram alterations \textit{\textbf{\textcolor{blue}{predict}}} ventricular fibrillation in Brugada syndrome. & \textit{\textbf{\textcolor{blue}{Do}}} spontaneous electrocardiogram alterations \textit{\textbf{\textcolor{blue}{predict}}} ventricular fibrillation in Brugada syndrome? & \textit{\textbf{\textcolor{blue}{yes}}} & 92.8 \\
\midrule
Liver grafts from selected older donors \textit{\textbf{\textcolor{red}{do not have}}} significantly more ischaemia reperfusion injury. & \textit{\textbf{\textcolor{red}{Do}}} liver grafts from selected older donors \textit{\textbf{\textcolor{red}{have}}} significantly more ischaemia reperfusion injury? & \textit{\textbf{\textcolor{red}{no}}} & 7.2 \\
\bottomrule
\end{tabular}
\caption{Examples of automatically generated instances for PQA-A. Original statement titles are converted to questions and answers are automatically generated according to the negation status.}
\label{tab:pqaa}
\vspace{-0.5em}
\end{table*}

\paragraph{Collection of PQA-A:} Motivated by the recent successes of large-scale pre-training from ELMo \cite{peters2018deep} and BERT \cite{devlin2018bert}, we use a simple heuristic to collect many noisily-labeled instances to build PQA-A for pre-training. Towards this end, we use PubMed articles with 1) a statement title which has POS tagging structures of NP-(VBP/VBZ)\footnote{Using Stanford CoreNLP parser \cite{manning-EtAl:2014:P14-5}.} and 2) a structured abstract including a conclusive part. The statement titles are converted to questions by simply moving or adding copulas (``is'', ``are'') or auxiliary verbs (``does'', ``do'') in the front and further revising for coherence (e.g.: adding a question mark). We generate the yes/no answer according to negation status of the VB. Several examples are shown in Table \ref{tab:pqaa}. We collected 211.3k instances for PQA-A, of which 200k randomly sampled instances are for training and the rest 11.3k instances are for validation.

\subsection{Characteristics}

We show the basic statistics of three PubMedQA subsets in Table \ref{tab:stat}.

\begin{figure}
    \centering
    \includegraphics[width=\linewidth]{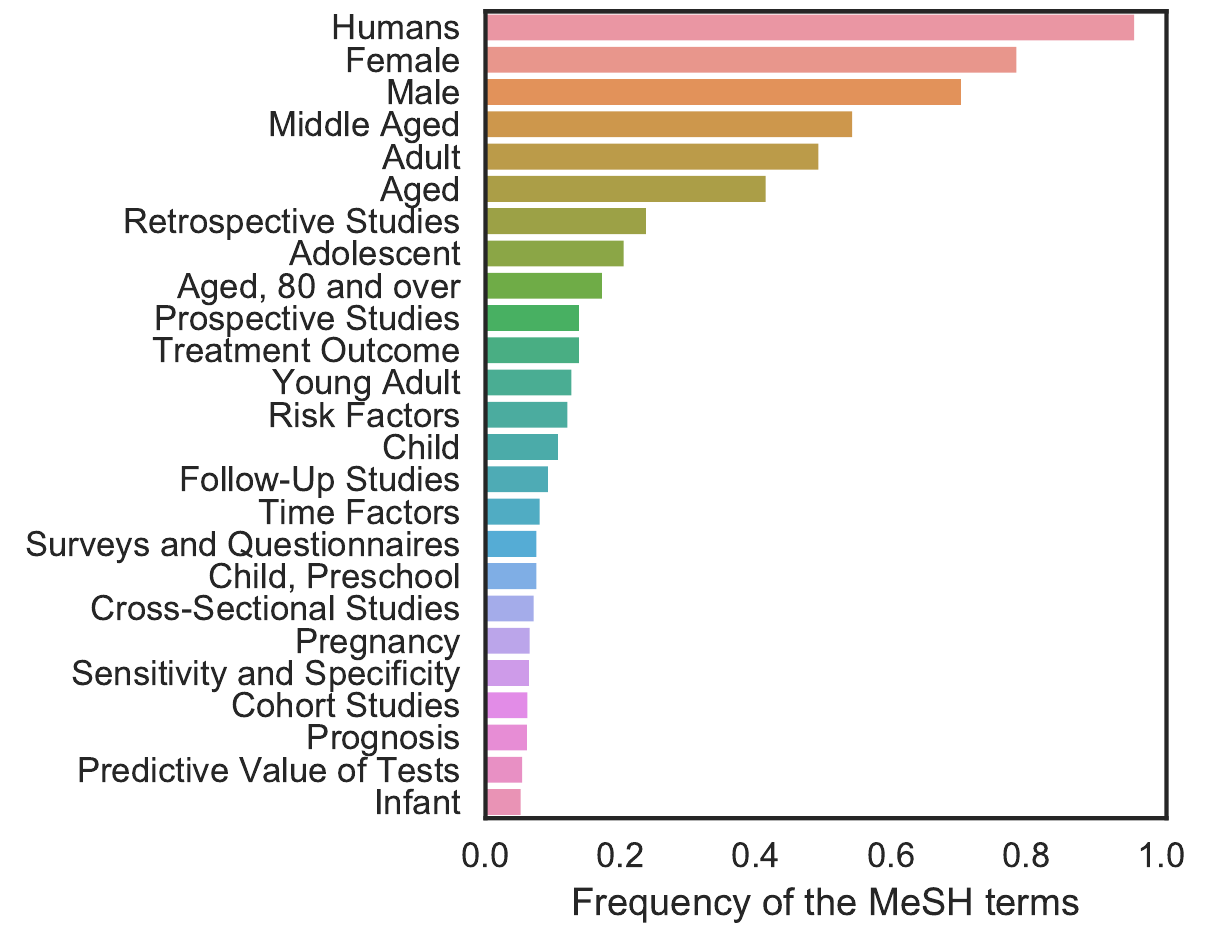}
    \vskip -0.15cm
    \caption{MeSH topic distribution of PubMedQA.}
    \label{fig:mesh}
    \vspace{-0.5em}
\end{figure}

\paragraph{Instance Topics:} PubMed abstracts are manually annotated by medical librarians with Medical Subject Headings (MeSH)\footnote{\url{https://www.nlm.nih.gov/mesh}}, which is a controlled vocabulary designed to describe the topics of biomedical texts. We use MeSH terms to represent abstract topics, and visualize their distribution in Fig. \ref{fig:mesh}. Nearly all instances are human studies and they cover a wide variety of topics, including retrospective, prospective, and cohort studies, different age groups, and healthcare-related subjects like treatment outcome, prognosis and risk factors of diseases.

\begin{table*}[!t]
    \centering
    \small
    \begin{tabular}{p{4.6cm}cp{9.5cm}}
        \toprule
        \textbf{Question Type} & \textbf{\%} & \textbf{Example Questions}  \\ 
        \midrule
        Does a factor \textbf{\textit{\textcolor{blue}{influence}}} the output? & 36.5 & Does reducing spasticity \textbf{\textit{\textcolor{blue}{translate into}}} functional benefit? \\
        & & Does ibuprofen \textbf{\textit{\textcolor{blue}{increase}}} perioperative blood loss during hip arthroplasty? \\
        \midrule
        Is a therapy \textbf{\textit{\textcolor{blue}{good/necessary}}}? & 26.0 & \textbf{\textit{\textcolor{blue}{Should}}} circumcision be performed in childhood? \\
        & & Is external palliative radiotherapy for gallbladder carcinoma \textbf{\textit{\textcolor{blue}{effective}}}? \\
        \midrule
        Is a \textbf{\textit{\textcolor{blue}{statement}}} true? & 18.0 & Sternal fracture in growing children: \textbf{\textit{\textcolor{blue}{A rare and often overlooked fracture?}}} \\
        & & Xanthogranulomatous cholecystitis: \textbf{\textit{\textcolor{blue}{a premalignant condition}}}? \\
        \midrule
        Is a factor \textbf{\textit{\textcolor{blue}{related to}}} the output? & 18.0 & Can PRISM \textbf{\textit{\textcolor{blue}{predict}}} length of PICU stay? \\
        & & Is trabecular bone \textbf{\textit{\textcolor{blue}{related to}}} primary stability of miniscrews? \\
        \toprule
        \textbf{Reasoning Type} & \textbf{\%} & \textbf{Example Snippet in Context}  \\
        \midrule
        \textit{\textbf{\textcolor{red}{Inter-group}}} comparison & 57.5 & [...] Postoperative AF was significantly lower in the \textit{\textbf{\textcolor{red}{Statin group}}} compared with the \textit{\textbf{\textcolor{red}{Non-statin group}}} (16\% versus 33\%, p=0.005). [...] \\
        \midrule
        Interpreting \textit{\textbf{\textcolor{red}{subgroup}}} statistics & 16.5 & [...] 57\% of patients were \textit{\textbf{\textcolor{red}{of lower socioeconomic status}}} and they had more health problems, less functioning, and more symptoms [...] \\
        \midrule
        {Interpreting \textit{\textbf{\textcolor{red}{(single) group}}} statistics} & 16.0 & [...] \textit{\textbf{\textcolor{red}{A total of 4 children}}} aged 5-14 years with a sternal fracture were treated in 2 years, 2 children were hospitalized for pain management and [...]\\
        \toprule
        \textbf{Text Interpretations of Numbers} & \textbf{\%} & \textbf{Example Snippet in Context} \\ 
        \midrule
        Existing \textbf{\textit{\textcolor{blue}{interpretations}}} of \textbf{\textit{\textcolor{blue}{numbers}}} & 75.5 & [...] Postoperative AF was \textbf{\textit{\textcolor{blue}{significantly lower}}} in the Statin group compared with the Non-statin group (\textbf{\textit{\textcolor{blue}{16\% versus 33\%, p=0.005}}}). [...] \\
        \midrule
        No interpretations (\textbf{\textit{\textcolor{blue}{numbers only}}}) & 21.0 & [...] 30-day mortality was \textbf{\textit{\textcolor{blue}{12.4\%}}} in those aged$<$70 years and \textbf{\textit{\textcolor{blue}{22\%}}} in those$>$70 years (\textbf{\textit{\textcolor{blue}{p$<$0.001}}}). [...] \\
        \midrule
        No numbers (\textbf{\textit{\textcolor{blue}{texts only}}}) & 3.5 &  [...] The halofantrine therapeutic dose group showed \textbf{\textit{\textcolor{blue}{loss and distortion of inner hair cells and inner phalangeal cells}}} [...] \\
        \bottomrule
    \end{tabular}
    \vskip -0.15cm
    \caption{Summary of PubMedQA question types, reasoning types and whether there are text descriptions of the statistics in context. Colored texts are matched key phrases (sentences) between types and examples.
    }
    \label{tab:type}
    \vspace{-0.5em}
\end{table*}

\begin{figure}
    \centering
    \includegraphics[width=\linewidth]{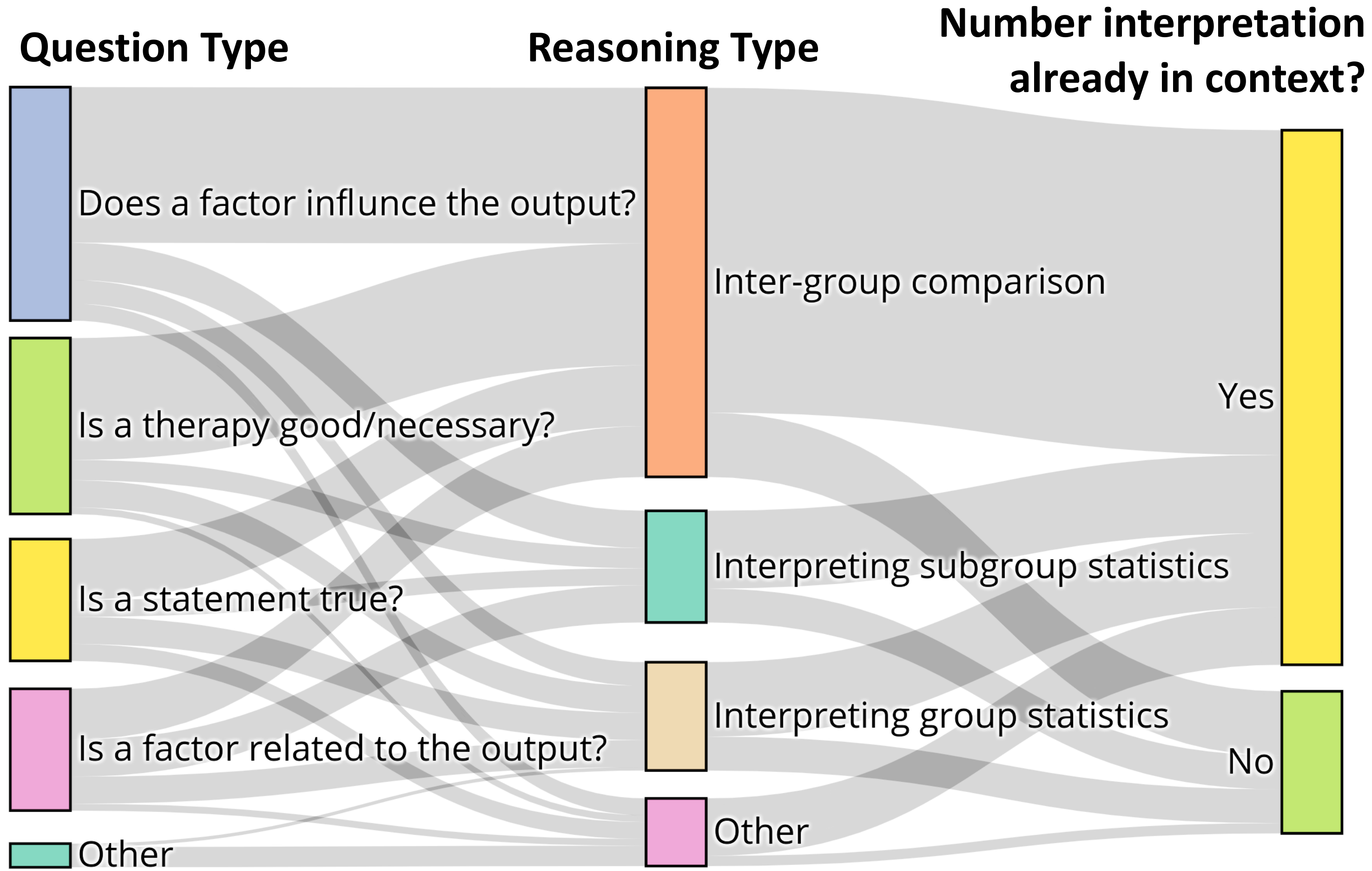}
    \vskip -0.15cm
    \caption{Proportional relationships between corresponded question types, reasoning types, and whether the text interpretations of numbers exist in contexts.}
    \label{fig:type}
\end{figure}

\paragraph{Question and Reasoning Types:} We sampled 200 examples from PQA-L and analyzed the types of questions and types of reasoning required to answer them, which is summarized in Table \ref{tab:type}. Various types of questions have been asked, including causal effects, evaluations of therapies, relatedness, and whether a statement is true. Besides, PubMedQA also covers several different reasoning types: most (57.5\%) involve comparing multiple groups (e.g.: experiment and control), and others require interpreting statistics of a single group or its subgroups. Reasoning over quantitative contents is required in nearly all (96.5\%) of them, which is expected due to the nature of biomedical research. 75.5\% of contexts have text descriptions of the statistics while 21.0\% only have the numbers. We use a Sankey diagram to show the proportional relationships between corresponded question type and reasoning type, as well as corresponded reasoning type and whether there are text interpretations of numbers in Fig. \ref{fig:type}.

\subsection{Evaluation Settings}
The main metrics of PubMedQA are accuracy and macro-F1 on PQA-L test set using question and context as input. We denote prediction using question and context as a \textbf{reasoning-required} setting, because under this setting answers are not directly expressed in the input and reasoning over the contexts is required to answer the question. Additionally, long answers are available at training time, so generation or prediction of them can be used as an auxiliary task in this setting.

A parallel setting, where models can use question and long answer to predict yes/no/maybe answer, is denoted as \textbf{reasoning-free} setting since yes/no/maybe are usually explicitly expressed in the long answers (i.e.: conclusions of the abstracts). Obviously, it's a much easier setting which can be exploited for bootstrapping PQA-U.

\section{Methods}

\subsection{Fine-tuning BioBERT}
We fine-tune BioBERT \cite{lee2019biobert} on PubMedQA as a baseline. BioBERT is initialized with BERT \cite{devlin2018bert} and further pre-trained on PubMed abstracts and PMC\footnote{\url{https://www.ncbi.nlm.nih.gov/pmc/}} articles. Expectedly, it vastly outperforms BERT in various biomedical NLP tasks. We denote the original transformer weights of BioBERT as $\theta_0$.

While fine-tuning, we feed PubMedQA questions and contexts (or long answers), separated by the special {\tt [SEP]} token, to BioBERT. The yes/no/maybe labels are predicted using the special {\tt [CLS]} embedding using a softmax function. Cross-entropy loss of predicted and true label distribution is denoted as $\mathcal{L}_\text{QA}$.

\subsection{Long Answer as Additional Supervision}
Under reasoning-required setting, long answers are available in training but not inference phase. We use them as an additional signal for training: similar to \citet{ma2018bag} regularizing neural machine translation models with binary bag-of-word (BoW) statistics, we fine-tune BioBERT with an auxiliary task of predicting the binary BoW statistics of the long answers, also using the special {\tt [CLS]} embedding. We minimize binary cross-entropy loss of this auxiliary task:
\[
\mathcal{L}_\text{BoW} = - \frac{1}{N} \sum_i b_i \text{log}\hat{b_i} + (1-b_i) \text{log}(1-\hat{b_i})
\]
where $b_i$ and $\hat{b_i}$ are ground-truth and predicted probability of whether token $i$ is in the long answers (i.e.: $b_i \in \{0, 1\}$ and $\hat{b_i} \in [0, 1]$), and $N$ is the BoW vocabulary size. The total loss is:
\[
\mathcal{L} = \mathcal{L}_\text{QA} + \beta \mathcal{L}_\text{BoW}
\]
In reasoning-free setting which we use for bootstrapping, the regularization coefficient $\beta$ is set to 0 because long answers are directly used as input.

\subsection{Multi-phase Fine-tuning Schedule} \label{notation}
\begin{figure}
    \centering
    \includegraphics[width=\linewidth]{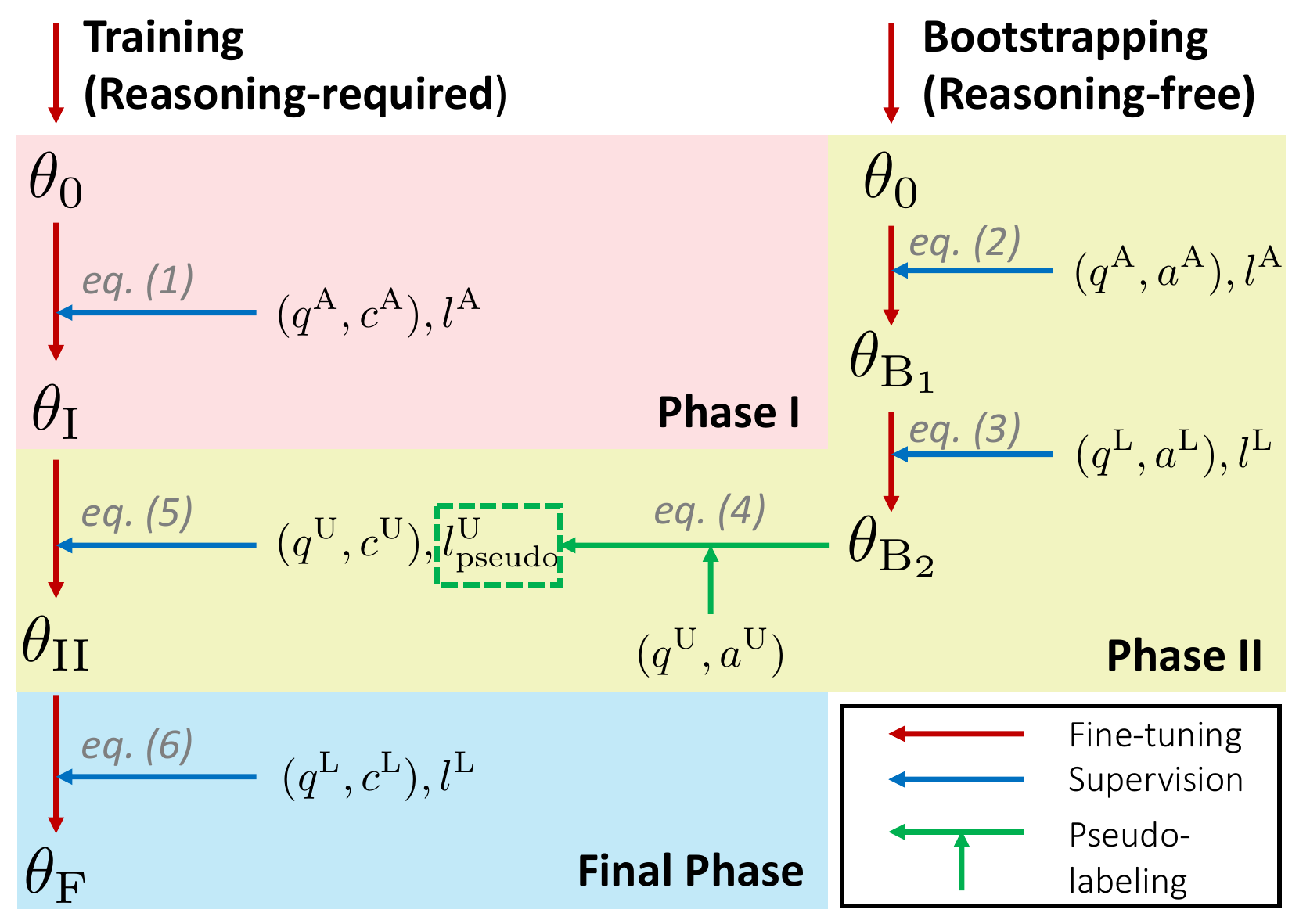}
    \vskip -0.15cm
    \caption{Multi-phase fine-tuning architecture. Notations and equations are described in \S \ref{notation}.}
    \label{fig:arch}
    \vspace{-0.5em}
\end{figure}

Since PQA-A and PQA-U have different properties from the ultimate test set of PQA-L, BioBERT is fine-tuned in a multi-phase style on different subsets. Fig. \ref{fig:arch} shows the architecture of this training schedule. We use $q$, $c$, $a$, $l$ to denote question, context, long answer and yes/no/maybe label of instances, respectively. Their source subsets are indexed by the superscripts of A for PQA-A, U for PQA-U and L for PQA-L.

\paragraph{Phase I Fine-tuning on PQA-A:}
PQA-A is automatically collected whose questions and labels are artificially generated. As a result, questions of PQA-A might differ a lot from those of PQA-U and PQA-L, and it only has yes/no labels with a very imbalanced distribution (92.8\% yes v.s. 7.2\% no). Despite these drawbacks, PQA-A has substantial training instances so models could still benefit from it as a pre-training step.

Thus, in Phase I of multi-phase fine-tuning, we initialize BioBERT with $\theta_0$, and fine-tune it on PQA-A using question and context as input:
\[ \label{eq1}
\theta_\text{I} \gets \mathrm{argmin}_\theta \, \mathcal{L}(\text{BioBERT}_{\theta}(q^\text{A},c^\text{A}), l^\text{A}) \tag{1}
\]

\paragraph{Phase II Fine-tuning on Bootstrapped PQA-U:}
To fully utilize the unlabeled instances in PQA-U, we exploit the easiness of reasoning-free setting to pseudo-label these instances with a bootstrapping strategy: first, we initialize BioBERT with $\theta_0$, and fine-tune it on PQA-A using question and long answer (reasoning-free), 
\[ \label{eq2}
\theta_{\text{B}_1} \gets \mathrm{argmin}_\theta \, \mathcal{L}(\text{BioBERT}_{\theta}(q^\text{A},a^\text{A}), l^\text{A}) \tag{2}
\]

then we further fine-tune $\text{BioBERT}_{\theta_{\text{B}_1}}$ on PQA-L, also under the reasoning-free setting:
\[ \label{eq3}
\theta_{\text{B}_2} \gets \mathrm{argmin}_\theta \, \mathcal{L}(\text{BioBERT}_{\theta}(q^\text{L},a^\text{L}), l^\text{L}) \tag{3}
\]

We pseudo-label PQA-U instances using the most confident predictions of $\text{BioBERT}_{\theta_{\text{B}_2}}$ for each class. Confidence is simply defined by the corresponding softmax probability and then we label a subset which has the same proportions of yes/no/maybe labels as those in the PQA-L:
\[\label{eq4}
l^\text{U}_\text{pseudo} \gets \text{BioBERT}_{\theta_{\text{B}_2}}(q^\text{U},a^\text{U}) \tag{4}
\]

In phase II, we fine-tune $\text{BioBERT}_{\theta_\text{I}}$ on the bootstrapped PQA-U using question and context (under reasoning-required setting):
\[ \label{eq5}
\theta_\text{II} \gets \mathrm{argmin}_\theta \, \mathcal{L}(\text{BioBERT}_{\theta}(q^\text{U},c^\text{U}), l^\text{U}_\text{pseudo}) \tag{5}
\]

\paragraph{Final Phase Fine-tuning on PQA-L:}
In the final phase, we fine-tune $\text{BioBERT}_{\theta_\text{II}}$ on PQA-L:
\[ \label{eq6}
\theta_{\text{F}} \gets \mathrm{argmin}_\theta \, \mathcal{L}(\text{BioBERT}_{\theta}(q^\text{L},c^\text{L}), l^\text{L}) \tag{6}
\]

Final predictions on instances of PQA-L validation and test sets are made using $\text{BioBERT}_{\theta_\text{F}}$:
\[
l_\text{pred} = \text{BioBERT}_{\theta_\text{F}}(q^\text{L},c^\text{L})
\]

\subsection{Compared Models}
\paragraph{Majority:} The majority (about 55\%) of the instances have the label ``yes''. We use a trivial baseline denoted as Majority where we simply predict ``yes'' for all instances, regardless of the question and context.

\paragraph{Shallow Features:} For each instance, we include the following shallow features: 1) TF-IDF statistics of the question 2) TF-IDF statistics of the context/long answer and 3) sum of IDF of the overlapping non-stop words between the question and the context/long answer. To allow multi-phase fine-tuning, we apply a feed-forward neural network on the shallow features instead of using a logistic classifier.

\paragraph{BiLSTM:} We simply concatenate the question and context/long answer with learnable segment embeddings appended to the biomedical word2vec embeddings \cite{Pyysalo2013DistributionalSR} of each token. The concatenated sentence is then fed to a biLSTM, and the final hidden states of the forward and backward network are used for classifying the yes/no/maybe label.

\paragraph{ESIM with BioELMo:} Following the state-of-the-art recurrent architecture of NLI \cite{peters2018deep}, we use pre-trained biomedical contextualized embeddings BioELMo \cite{jin2019probing} for word representations. Then we apply the ESIM model \cite{chen2016enhanced}, where a biLSTM is used to encode the question and context/long answer, followed by an attentional local inference layer and a biLSTM inference composition layer. After pooling, a softmax output unit is applied for predicting the yes/no/maybe label.

\subsection{Compared Training Schedules}
\paragraph{Final Phase Only:}
Under this setting, we train models only on PQA-L. It's an extremely low resources setting where there are only 450 training instances in each fold of cross-validation.

\paragraph{Phase I + Final Phase:}
Under this setting, we skip the training on bootstrapped PQA-U. Models are first fine-tuned on PQA-A, and then fine-tuned on PQA-L.

\paragraph{Phase II + Final Phase:}
Under this setting, we skip the training on PQA-A. Models are first fine-tuned on bootstrapped PQA-U, and then fine-tuned on PQA-L. 

\paragraph{Single-phase Training:}
Instead of training a model sequentially on different splits, under single-phase training setting we train the model on the combined training set of all PQA splits: PQA-A, bootstrapped PQA-U and PQA-L.

\section{Experiments}

\subsection{Human Performance} \label{human}
Human performance is measured during the annotation: As shown in Algorithm \ref{algo:pqal}, annotations of annotator 1 and annotator 2 are used to calculate reasoning-free and reasoning-required human performance, respectively, against the discussed ground truth labels. Human performance on the test set of PQA-L is shown in Table \ref{tab:human}. We only test single-annotator performance due to limited resources. \citet{kwiatkowski2019natural} show that an ensemble of annotators perform significantly better than single-annotator, so the results reported in Table \ref{tab:human} are the lower bounds of human performance. Under reasoning-free setting where the annotator can see the conclusions, a single human achieves 90.4\% accuracy and 84.2\% macro-F1. Under reasoning-required setting, the task becomes much harder, but it's still possible for humans to solve: a single annotator can get 78.0\% accuracy and 72.2\% macro-F1.
\begin{table}[htbp]
\centering
\small
\begin{tabular}{lcc}
\toprule
\textbf{Setting} & \textbf{Accuracy (\%)} & \textbf{Macro-F1 (\%)} \\
\midrule
Reasoning-Free & 90.40 & 84.18 \\
Reasoning-Required & 78.00 & 72.19 \\
\bottomrule
\end{tabular}
\vskip -0.15cm
\caption{Human performance (single-annotator).}
\label{tab:human}
\end{table}

\subsection{Main Results}
\begin{table*}
    \small
    \centering
    \begin{tabular}{lcccccccccc}
        \toprule
        \multirow{2}{*}{\textbf{Model}} & 
        \multicolumn{2}{c}{\textbf{Final Phase Only}} &
        \multicolumn{2}{c}{\textbf{Single-phase}} &
        \multicolumn{2}{c}{\textbf{Phase I + Final}} &
        \multicolumn{2}{c}{\textbf{Phase II + Final}} & 
        \multicolumn{2}{c}{\textbf{Multi-phase}} \\
        \cmidrule{2-11}
        & \textbf{Acc} & \textbf{F1} & \textbf{Acc} & \textbf{F1} & \textbf{Acc} & \textbf{F1} & \textbf{Acc} & \textbf{F1} & \textbf{Acc} & \textbf{F1} \\
        \midrule
        Majority & 55.20 & 23.71 & -- & -- & -- & -- & -- & -- & -- & -- \\
        Human (single) & 78.00 & 72.19 & -- & -- & -- & -- & -- & -- & -- & -- \\
        \midrule
        w/o A.S. \\
        \midrule
        \hspace{0.1cm}Shallow Features & 53.88 & 36.12 & \underline{57.58} & 31.47 & 57.48 & 37.24 & 56.28 & \underline{40.88} & 53.50 & 39.33 \\
        \hspace{0.1cm}BiLSTM & 55.16 & 23.97 & 55.46 & 39.70 & 58.44 & 40.67 & 52.98 & 33.84 & 59.82 & \underline{41.86} \\
        \hspace{0.1cm}ESIM w/ BioELMo & 53.90 & 32.40 & 61.28 & 42.99 & 61.96 & 43.32 & 60.34 & 44.38 & 62.08 & 45.75 \\
        \hspace{0.1cm}BioBERT & 56.98 & 28.50  & 66.44 & 47.25 & 66.90 & 46.16 & 66.08 & 50.84 & 67.66 & 52.41 \\
        \midrule
        w/ A.S. \\
        \midrule
        \hspace{0.1cm}Shallow Features & 53.60 & 35.92 & 57.30 & 30.45 & 55.82 & 35.09 & 56.46$^\dagger$ & 40.76 & 55.06$^\dagger$ & 40.67$^\dagger$ \\
        \hspace{0.1cm}BiLSTM & 55.22$^\dagger$ & 23.86 & 55.96$^\dagger$ & 40.26$^\dagger$ & \underline{61.06}$^\dagger$ & 41.18$^\dagger$ & 54.12$^\dagger$ & 34.11$^\dagger$ & 58.86 & 41.06 \\
        \hspace{0.1cm}ESIM w/ BioELMo & 53.96$^\dagger$ & 31.07 & 62.68$^\dagger$ & 43.59$^\dagger$ & \underline{63.72}$^\dagger$ & 47.04$^\dagger$ & 60.16 & 45.81$^\dagger$ & \underline{63.72}$^\dagger$ & \underline{47.90}$^\dagger$ \\
        \hspace{0.1cm}BioBERT & 57.28$^\dagger$ & 28.70$^\dagger$ & 66.66$^\dagger$ & 46.70$^\dagger$ & 67.24$^\dagger$ & 46.21$^\dagger$ & 66.44$^\dagger$ & 51.41$^\dagger$ & \textbf{\underline{68.08}}$^\dagger$ & \textbf{\underline{52.72}}$^\dagger$ \\
        \bottomrule
    \end{tabular}
    \vskip -0.15cm
    \caption{Main results on PQA-L test set under reasoning-required setting. A.S.: additional supervision. $^\dagger$with A.S. is better than without A.S. Underlined numbers are model-wise best performance, and bolded numbers are global best performance. All numbers are percentages.} \label{tab:main}
\end{table*}
We report the test set performance of different models and training schedules in Table \ref{tab:main}. In general, multi-phase fine-tuning of BioBERT with additional supervision outperforms other baselines by large margins, but the results are still much worse than just single-human performance. 

\paragraph{Comparison of Models:} A trend of BioBERT $>$ ESIM w/ BioELMo $>$ BiLSTM $>$ shallow features $>$ majority, conserves across different training schedules on both accuracy and macro-F1. Fine-tuned BioBERT is better than state-of-the-art recurrent model of ESIM w/ BioELMo, probably because BioELMo weights are fixed while all BioBERT parameters can be fine-tuned, which better benefit from the pre-training settings.

\paragraph{Comparison of Training Schedules:} Multi-phase fine-tuning setting gets 5 out of 9 model-wise best accuracy/macro-F1. Due to lack of annotated data, training only on the PQA-L (final phase only) generates similar results as the majority baseline. In phase I + Final setting where models are pre-trained on PQA-A, we observe significant improvements on accuracy and macro-F1 and some models even achieve their best accuracy under this setting. This indicates that a hard task with limited training instances can be at least partially solved by pre-training on a large automatically collected dataset when the tasks are similarly formatted.

Improvements are also observed in phase II + Final setting, though less significant than those of phase I + Final. As expected, multi-phase fine-tuning schedule is better than single-phase, due to different properties of the subsets.

\paragraph{Additional Supervision:} Despite its simplicity, the auxiliary task of long answer BoW prediction clearly improves the performance: most results (28/40) are better with such additional supervision than without.

\subsection{Intermediate Results}
In this section we show the intermediate results of multi-phase fine-tuning schedule.

\begin{table}
    \small
    \centering
    \begin{tabular}{lcccc}
        \toprule
        \multirow{2}{*}{\textbf{Model}} & 
        \multicolumn{2}{c}{\textbf{w/o A.S.}} &
        \multicolumn{2}{c}{\textbf{w/ A.S.}} \\
        \cmidrule{2-5}
        & \textbf{Acc} & \textbf{F1} & \textbf{Acc} & \textbf{F1} \\
        \midrule
        Majority & 92.76 & 48.12 & -- & -- \\
        \midrule
        Shallow Features & 93.01 & 54.59 & 93.05 & 55.12  \\
        BiLSTM & 94.59 & 73.40 & 94.45 & 71.81 \\
        ESIM w/ BioELMo & 94.82 & 74.01 & 95.04 & 75.22 \\
        BioBERT & 96.50 & 84.65 & 96.40 & 83.76\\
        \bottomrule
    \end{tabular}
    \vskip -0.15cm
    \caption{Results of Phase I (eq. \ref{eq1}). Experiments are on PQA-A under reasoning-required setting. A.S.: additional supervision.} 
    \label{tab:phasei}
\end{table}

\begin{table}
    \small
    \centering
    \begin{tabular}{lcccc}
        \toprule
        \multirow{2}{*}{\textbf{Model}} & 
        \multicolumn{2}{c}{\textbf{Eq. \ref{eq2}}} &
        \multicolumn{2}{c}{\textbf{Eq. \ref{eq3}}} \\
        \cmidrule{2-5}
        & \textbf{Acc} & \textbf{F1} & \textbf{Acc} & \textbf{F1} \\
        \midrule
        Majority & 92.76 & 48.12 & 55.20 & 23.71 \\
        Human (single) & -- & -- & 90.40$^\dagger$ & 84.18$^\dagger$ \\
        \midrule
        Shallow Features & 93.11 & 56.11 & 54.44 & 38.63 \\
        BiLSTM & 95.97 & 83.70 & 71.46 & 50.93 \\
        ESIM w/ BioELMo & 97.01 & 88.47 & 74.06 & 58.53 \\
        BioBERT & {98.28} & {93.17} & {80.80} & {63.50} \\
        \bottomrule
    \end{tabular}
    \vskip -0.15cm
    \caption{Bootstrapping results. Experiments are on PQA-A (eq. \ref{eq2}) and PQA-L (eq. \ref{eq3}) under reasoning-free setting. $^\dagger$Reasoning-free human performance.} 
    \label{tab:bootstrap}
\end{table}

\begin{table}
    \small
    \centering
    \begin{tabular}{lcccc}
        \toprule
        \multirow{2}{*}{\textbf{Model}} & 
        \multicolumn{2}{c}{\textbf{w/o A.S.}} &
        \multicolumn{2}{c}{\textbf{w/ A.S.}} \\
        \cmidrule{2-5}
        & \textbf{Acc} & \textbf{F1} & \textbf{Acc} & \textbf{F1} \\
        \midrule
        Majority & 55.10 & 23.68 & -- & -- \\
        Shallow Features & 76.66 & 66.12 & 77.71 & 67.97 \\
        \midrule
        Majority & 56.53 & 24.07 & -- & -- \\
        BiLSTM & 85.33 & 81.32 & 85.68 & 81.87 \\
        \midrule
        Majority & 55.10 & 23.68 & -- & -- \\
        ESIM w/ BioELMo & 78.47 & 63.32 & 79.62 & 64.91 \\
        \midrule
        Majority & 54.82 & 24.87 & -- & -- \\        
        BioBERT & 80.93 & 68.84 & 81.02 & 70.04 \\
        \bottomrule
    \end{tabular}
    \vskip -0.15cm
    \caption{Phase II results (eq. \ref{eq5}). Experiments are on pseudo-labeled PQA-U under reasoning-required setting. A.S.: additional supervision.
    }    
    \label{tab:phaseii}
\end{table}

\paragraph{Phase I:} Results are shown in Table \ref{tab:phasei}. Phase I is fine-tuning on PQA-A using question and context. Since PQA-A is imbalanced due to its collection process, a trivial majority baseline gets 92.76\% accuracy. Other models have better accuracy and especially macro-F1 than majority baseline. Fine-tuned BioBERT performs best. 

\paragraph{Bootstrapping:} Results are shown in Table \ref{tab:bootstrap}. Bootstrapping is a three-step process: fine-tuning on PQA-A, then on PQA-L and pseudo-labeling PQA-U. All three steps are using question and long answer as input. Expectedly, models perform better in this reasoning-free setting than they do in reasoning-required setting (for PQA-A, Eq. 2 results in Table \ref{tab:bootstrap} are better than the performance in Table \ref{tab:phasei}; for PQA-L, Eq. 3 results in Table \ref{tab:bootstrap} are better than the performance in Table \ref{tab:main}).

\paragraph{Phase II:} Results are shown in Table \ref{tab:phaseii}. In Phase II, since each model is fine-tuned on its own pseudo-labeled PQA-U instances, results are not comparable between models. While the ablation study in Table \ref{tab:main} clearly shows that Phase II is helpful, performance in Phase II doesn't necessarily correlate with final performance on PQA-L. 

\section{Conclusion}
We present PubMedQA, a novel dataset aimed at biomedical research question answering using yes/no/maybe, where complex quantitative reasoning is required to solve the task. PubMedQA has substantial automatically collected instances as well as the largest size of expert annotated yes/no/maybe questions in biomedical domain. We provide a strong baseline using multi-phase fine-tuning of BioBERT with long answer as additional supervision, but it's still much worse than just single human performance. 

There are several interesting future directions to explore on PubMedQA, e.g.: (1) about 21\% of PubMedQA contexts contain no natural language descriptions of numbers, so how to properly handle these numbers is worth studying; (2)  we use binary BoW statistics prediction as a simple demonstration for additional supervision of long answers. Learning a harder but more informative auxiliary task of long answer generation might lead to further improvements.

Articles of PubMedQA are biased towards clinical study-related topics (described in Appendix \ref{appendix:topic}), so PubMedQA has the potential to assist evidence-based medicine, which seeks to make clinical decisions based on evidence of high quality clinical studies. Generally, PubMedQA can serve as a benchmark for testing scientific reasoning abilities of machine reading comprehension models.

\section{Acknowledgement}
We are grateful for the anonymous reviewers of EMNLP who gave us very valuable comments and suggestions.

\bibliography{emnlp-ijcnlp-2019}
\bibliographystyle{acl_natbib}

\clearpage
\appendix

\section{Yes/no/maybe Answerability}
Not all naturally occuring question titles from PubMed are answerable by yes/no/maybe. The first step of annotating PQA-L (as shown in algorithm \ref{algo:pqal}) from pre-PQA-U is to manually identify questions that can be answered using yes/no/maybe. We labeled 1091 (about 50.2\%) of 2173 question titles as unanswerable. For example, those questions cannot be answered by yes/no/maybe:
\begin{itemize}
    \item ``Critical Overview of HER2 Assessement in Bladder Cancer: What Is Missing for a Better Therapeutic Approach?" (wh- question)
    \item ``Otolaryngology externships and the match: Productive or futile?" (multiple choices)
\end{itemize}

\section{Over-represented Topics} \label{appendix:topic}
Clinical study-related topics are over-represented in PubMedQA: we found proportions of MeSH terms like:
\begin{itemize}
    \item ``Pregnancy Outcome"
    \item ``Socioeconomic Factors"
    \item ``Risk Assessment"
    \item ``Survival Analysis"
    \item ``Prospective Studies"
    \item ``Case-Control Studies"
    \item ``Reference Values"
\end{itemize}
are significantly higher in the PubMedQA articles than those in 200k most recent general PubMed articles (significance is defined by $p<0.05$ in two-proportion z-test).

\section{Annotation Criteria}
Strictly speaking, most yes/no/maybe research questions can be answered by ``maybe" since there will always be some conditions where one statement is true and vice versa. However, the task will be trivial in this case. Instead, we annotate a question using ``yes" if the experiments and results in the paper indicate it, so the answer is not universal but context-dependent.

Given a question like ``Do patients benefit from drug X?": certainly not all patients will benefit from it, but if there is a significant difference in an outcome between the experimental and control group, the answer will be ``yes". If there is not, the answer will be ``no".

``Maybe" is annotated when (1) the paper discusses conditions where the answer is True and conditions where the answer is False or (2) more than one intervention/observation/etc. is asked, and the answer is True for some but False for the others (e.g.: ``Do Disease A, Disease B and/or Disease C benefit from drug X?"). To model uncertainty of the answer, we don't strictly follow the logic calculations where such questions can always be answered by either ``yes" or ``no".

\end{document}